# AI Knows Which Words Will Appear in Next Year's Korean CSAT

Byunghyun Ban*
AIEng.
https://aieng.kr
Anong-si, Republic of Korea
bhban@kakao.com

Jejong Lee
AIEng.
https://aieng.kr
Andong-si, Republic of Korea
aa5025@naver.com

Hyeonmok Hwang
AIEng.
https://aieng.kr
Yeongyang-gun, Republic of Korea
sbcris11@gmail.com

*Abstract*—A text-mining-based word class categorization method and LSTM-based vocabulary pattern prediction method are introduced in this paper. A preprocessing method based on simple text appearance frequency analysis is first described. This method was developed as a data screening tool but showed 4.35 times higher prediction accuracy compared to Word Master book. An LSTM deep learning method is also suggested for vocabulary appearance pattern prediction method. AI performs a regression with various size of data window of previous exams to predict the probabilities of word appearance in the next exam. Predicted values of AI over various data windows are processed into a single score as a weighted sum, which we call an "AI-Score", which represents the probability of word appearance in next year's exam. Suggested method showed 100% accuracy at the range 100-score area and showed only 1.7% error of prediction in the section where the scores were over 60 points. All source codes are freely available at the authors' Git Hub repository.
(https://github.com/needleworm/bigdata_voca)

*Keywords—Natural Language Processing, NLP, Text Mining, Deep Learning, Pattern Prediction*

## I. Introduction

Korean CSAT(대학수학능력시험) is an annual test administered by Korean Ministry of Education. Every Korean student takes K-CSAT for admission of university. As K-CSAT is held only once a year, Korea Institute of Curriculum and Evaluation (KICE) provides official mock exams in June and September. On average, 510 thousand students take K-CSAT every year [1]. As previous questions from K-CSAT are freely available at KICE, many instructors and educational institutes struggle for analyzing the pattern itself and the change of patterns over year from previous exams.

This paper provides a prediction matehod for english vocabulary appearance pattern with big-data mining and deep learning.

Previous works related to exam pattern analysis and natural language processing are summarized on II. Related works.

Details on data acquisition, preprocessing, screening and AI-based prediction are described in III. Method. It also provides detailed training strategies such as data windowing and loss function, and normalization method to convert AI-predicted logit into a score which measures the probability of word appearance.

IV. Results. shows the distribution and accuracy of AI-predicted values, and the prediction accuracy on 2023 K-CSAT English exam.

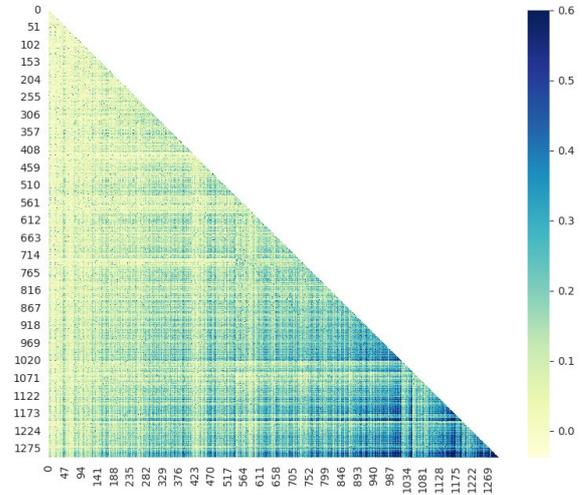

Fig 1. Heatmap for Word Correlation in Sentences

### A. Patterns of Exams

The author of Word Master introduces that he gathered English words from K-CSAT, official mock exam of K-CSAT, government authorized textbooks to write the vocabulary book [2]. Seonsik Kyeong claims that his book, which contains 4,138 words, predicted English vocabulary of K-CSAT with 93.3% ~ 98.5% accuracy for recent 3 years [3]. But each year's exam contained less than 1,600 words so maximum possible accuracy cannot numerically exceed 38.7%, which is achieved only when every word from an exam is contained in the book. It seems that the author did not apply mathematical definition: number of correct predictions over total number of predictions, or the result is not accurate.

Those two books are most popular in Korean English education book market, but not any statistical analysis results or big data analysis method about them are available.

Hyerang Lim provided a prediction method of difficulty of Korean CSAT English test, based on corpus analysis [4]. She applied Lexical Complexity Analyzer and L2 Syntactic Complexity Analyzer to investigate the relationship between linguistic features and item difficulty of a Mock CSAT English test. Y. Wang, et al. [5] and C. Zhang, et al. [6] proposed a neural network-based model with Long Short-Term Memory(LSTM) for SAT problem solving. Their LSTM model learns the question-answering pattern from SAT exam to perform problem solving.

Recent studies even provide various generative methods for educational-purposed question generation with artificial neural networks [6-9]. As even various generative model for question-answering is available, pattern analysis of a serial exam is not difficult at all.

### B. Natural Language Processing

Natural Language Processing (NLP) is a computer-science area for understanding and generating the patterns of human-speaking languages. Although linguistic definition of natural language is limited as naturally formed language along a community, NLP technologies can be also applied to artificial languages such as Esperanto or programming languages.

NLP approaches generally use linear algebraic approaches. So, most approaches for text NLP first convert the text or spoken words into tensor of integers. One of the easiest but most effective way is word vector embedding. Word vector embedding first requires a table of words to convert each word into a number. Whole vocabularies for analysis should be enlisted on the table. After writing the table, input sentences are tokenized and converted into a series of number, referencing the table. This method is called word2vec [10]. A sparse representation of word vectors, which just convert the word embedding vector into a set of one-hot vectors, can be applied for easier classification of each word because the one-hot vector representation is described as 2D data array which fits best for linear algebraic methods and neural networks.

Long Short-Term Memory(LSTM) was proposed by S.Hochreiter in 1997. [11] As LSTM has dynamic forget gate architecture, it has great advantage in learning time-series data and languages. Although transformer or CNN methods for NLP are highly recognized today, LSTM based NLP models are still the most popular algorithms [12].

## II. METHOD

The authors solemnly declare that this research is never intended to provide certain participant of Korean CSAT exam with an unfair advantage over other competitors. We wanted to rediscover and point out the positive roles of big data mining and artificial intelligence in English education. Also, we would like to suggest a method to predict vocabulary appearance pattern from short-scaled time series data with combination of AI-predicted probabilities inferred from various length of time scopes.

### A. Data

KICE's website (https://www.kice.re.kr) provides K-CSAT exam PDF files and the official mock test files. We obtained K-CSAT test files from 2003 to 2022, and mock test from 2004 Jun to 2022 September. As there were two types of English exam in 2014, total 21 PDF files of K-CSAT English exam and 40 mock test files were downloaded.

We used K-CSAT exam files for main analysis and the training resources of AI, while we used the mock test files for optimization of our method but not for training the neural networks.

### B. Preprocessing

All PDF files were converted into txt files with Python codes (Supplement Data 1). We have removed all Korean words and escape characters. We also replaced any useless patterns, Unicode errors, and some string patterns which appear only on some specific year's exam. 20 samples of removed tokens are described on Table 1.

TABLE I. 20 SAMPLES OF REMOVED TOKENS

| "\U000f0802" | "\n.\n" | "(A)," | "fffbg.\n" |
|---|---|---|---|
| "\U000f003b" | "., " | "(B)," | "(e.\ng" |
| "\x0c" | """" | "(C)," | "a.\nm." |
| "\xa0" | "________" | "(D)," | "p.\nm." |
| "/return" | "–" | "(E)," | """" |

Preprocessing rules are indicated on the *read_sunung_txts()* function of our source code (Supplement Data 2). After preprocessing, we also extracted sentences for analysis of word correlation in sentences. Then the processed texts are tokenized for frequency analysis.

### C. Frequency Analysis

6,575 words have appeared in K-CSAT, and 6,951 words have appeared in the mock test. 2,262 words have appeared in the mock test but never have appeared in the official K-CSAT. Also 1,886 words have only appeared in the official K-CSAT but never in the mock test. 4,689 words have appeared in both exams. The result shows that vocabulary tendency of the two tests is different.

However, word tokens were not directly used for pattern analysis because raw tokens contain not only the original forms of words but also plural forms of nouns, third person singular form of verbs, past tense form of verbs, and past or present participle form of verbs, etc. Therefore another step for restoring the original forms of word tokens were required. So we have built a dictionary which maps the words into their original forms with Selenium and Chrome Driver to convert the word tokens into their original form. Then the appearance frequency of each word was calculated along years.

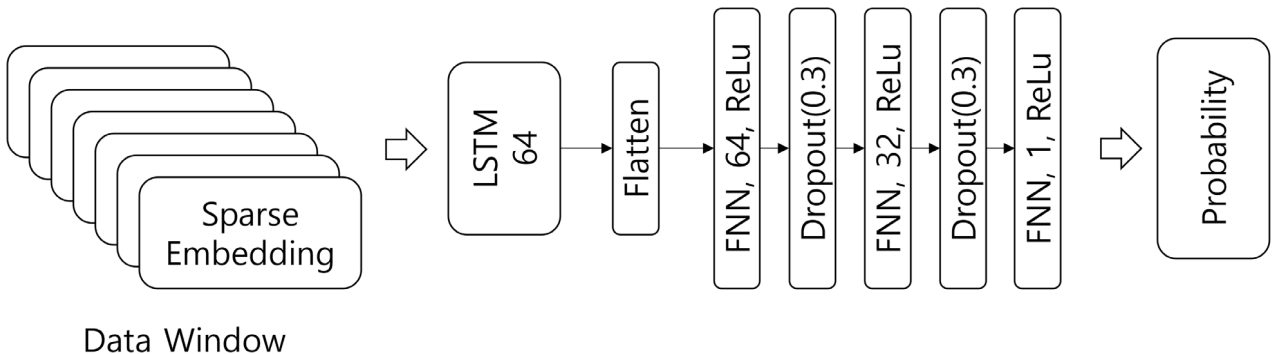

Fig 2. Neural Network Architecture

*D. Word Screening by Experts*

It is not so recommended to sort whole words by the frequencies of appearance because the grammatically repeated words, articles, pronouns, "to" for to infinitive, and many other tokens should be highly-ranked on the list. To analyze vocabulary tendency of time-series document data, we first had to remove some words which just technically have high-appearance-frequencies but do not provide patterns underlying sequential K-CSAT exam.

To remove such artifact for higher-resolution of analysis, two English education experts participated in the screening work; one was a public education teacher and the other was a private education teacher. They have screened proper nouns, inappropriate tokens, conjugations of verbs which computer could not process, and other words which do not fully reflect K-CSAT exam tendencies. After screening process, only 2,500 words were survived. The words were then categorized into two groups by frequencies of appearance.

Word group of 'High rank' contained 1,300 words, and the other group had 1,200 words. Finally, we used the 1,300 word group for training of AI.

The correlation of tokens of 'High rank' group in sentences are described in Fig 1. It shows that only small rate of words shows high correlation. We consider this result shows the removal of artifacts, because the result shows sparse correlation between words, which implies generalization of source data.

The frequency analysis result for 9,329 tokenized words is available at the authors' repository (Supplement Data 4). And 1,300 words' data over 19 years are available (Supplement Data 5).

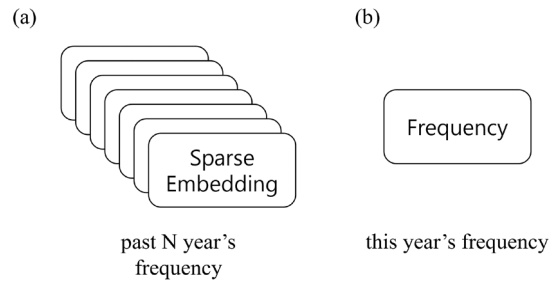

Fig 3. Data Windowing Strategy

*E. Neural Network Architecture*

We designed a small deep neural network model because the time-scale dimension of data was too small. As K-CSAT exam has only 20 years of history, a big model may show severe overfitting.

The whole architecture is displayed on Fig 2. The model has single LSTM layer and 3 FNN layers. Flatten layer is applied to convert the shape of LSTM output into lower dimension tensor. First 2 FNN layers has dropout with rate 0.3. The output layer does not have a softMax or other nonlinear activation function because the post-processing step just can normalize the positive values of output logits. ReLu was applied to remove negative values because negative-signed probability is not needed to predict word appearance pattern. The source code is uploaded online (Supplement Data 3).

*F. Data Windowing*

In order to learn the previous exam and to predict the appearance probability of future exam, we used data windowing strategy. Windowing strategy is described in Fig 3. Each window consists of past N year's frequency tensor data. As a frequency distribution of 1,300 words in each year is a vector, 1D data, the elongation of data on the perpendicular direction forms 2D matrix.

A minibatch for training obtains an array of several windows, therefore, the input LSTM layer of the neural network receives a (1300, 1, N, B) size tensor, where B is the size of minibatch. Or it can be also represented into (1300, N, B) by squeezing the unnecessary dimension.

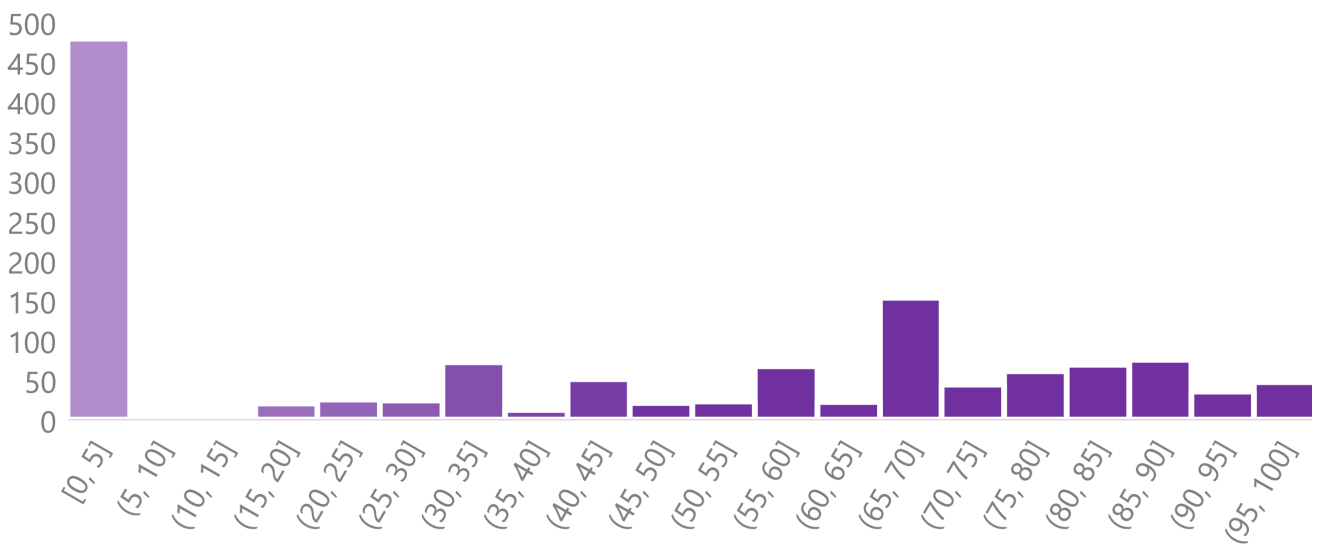

Fig 4. Distribution of AI-Scores.

*G. Training*

The model is then trained with the loss function described in equation (1). $X_{t-N,t}$ describes the data window of size N between t-N and t, and $X_{t+1}$ is the future probability at time t+1.

$$Loss(\theta) = \ln(\cosh((P_\theta(X_{t-N,t}) - X_{t+1})))  \quad (1)$$

It's a very simple strategy: with the given past tendency, predicting next year's word appearance tendency. We applied Adam optimizer [12] for stochastic weight readjustment.

We used TensorFlow 2.9.1 for implementation. And a MacBook Air with Apple Silicon(M1) was used for training.

*H. Normalization of Regression Results*

We trained the same model with 5 different-sized windows. After training the model with K-CSAT between 2003~2022, the trained model predicted frequency at 2023 exam, which was going to be held on November 17, 2022.

The predicted values for 5 different windows are described as 5 separated vectors with 1,300 values. Each value from different window was merged into single value by simple weighted sum. Window sizes used for the experiment and the weights are described on Table II.

TABLE II.    WEIGHTED SUM INFORMATION

| Window Size | Weight | Mean. Output. |
|---|---|---|
| 3 | 0.5 | 13.14 |
| 5 | 0.4 | 15.07 |
| 7 | 0.3 | 15.86 |
| 10 | 0.2 | 15.32 |
| 13 | 0.1 | 14.07 |

As larger window values implies long-ranged timeseries tendencies while smaller windows more concentrate on recent exams, we provided smaller weights for larger windows and bigger weights for smaller windows. We call the processed weighted sum as AI-Score.

## III. RESULT

*A. Bigdata Screening Method*

The frequency analysis result for 9,329 tokenized words is available at the authors' repository (Supplement Data 4). And 1,300 words' data over 19 years are available here (Supplement Data 5).

*B. AI Prediction Values*

Processed weighted sum of AI predicted values is available at ResearchGate (Supplement Data 6) and the distribution is described on Fig 4. The Model predicted 479 words will not appear on 2023 K-CSAT Exam.

*C. Prediction Result Compared to 2023 K-CSAT Exam*

After 2023's K-CSAT exam was held on November 17, 2022, we obtained and processed the English exam PDF file into processed tokens. Total 1,558 tokens including conjugations of verbs were extracted. After removing the duplicated words and conjugated word forms, 1,120 meaningful words remained. The list of words is available online (Supplement Data 7).

The prediction results of proposed methods and previous works are summarized on Table II. Our method showed 4.35 times higher accuracy compared to Previous Work 1, and 6.2 times higher accuracy compared to Previous Work 2.

Also, the rate of intersection between the set of words from 2023 K-CSAT and the set of the words in interest of the proposed method is 1.98 times higher than Previous Work 1, and 2.32 times higher than Previous Work 2.

Only the list of words from 2023 Korean CSAT English exams and with appearance data from proposed method and previous works are also provided on Supplement data 7.

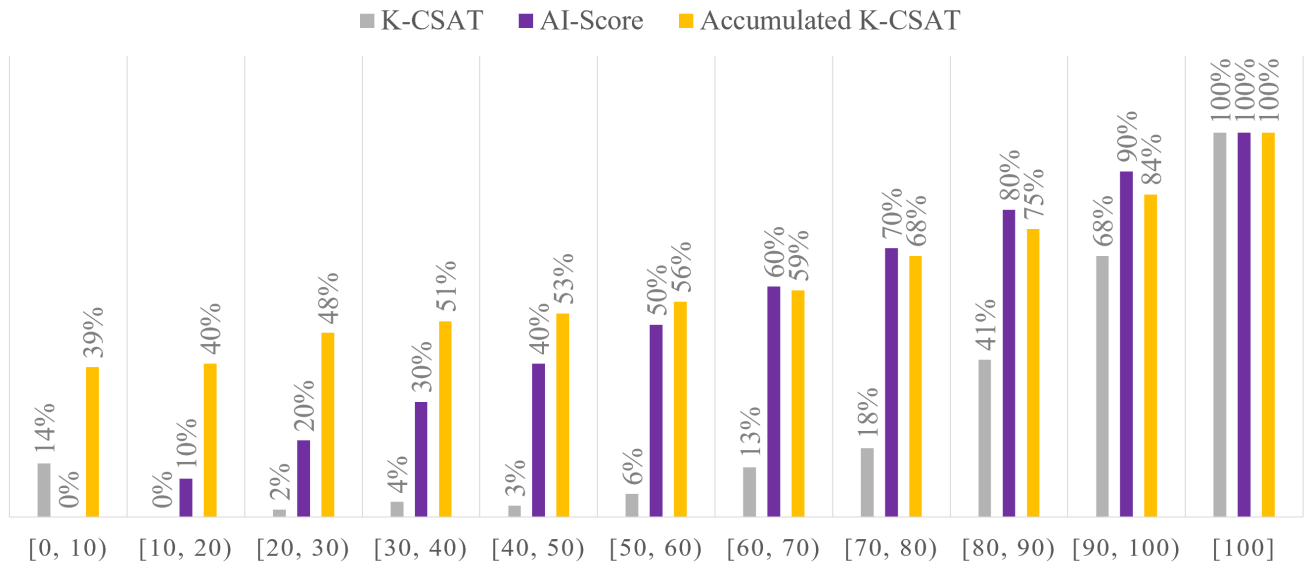

Fig 5.  Illustration for the accuracy of AI-predicted Scores.

TABLE III. PREDICTION RESULTS

|  | Proposed Method | Previous Work 1 [1] | *Previous Work 2 [2, 3]* |
| --- | --- | --- | --- |
| Predicted words* | 1,300 | 3,208 | 4,024 |
| Correct Prediction | 460 | 385 | 337 |
| Accuracy** | 52.2% | 12.0% | 8.4% |
| Intersection *** | 68.2% | 34.4% | 30.1% |

\* The number of words in interest (except duplicated words)
\*\* (# of True Positive) / (# of words in interest)
\*\*\* (# of True Positive) / (# of words from 2023 K-CSAT)

Although Copyright law is limited on research purpose so providing word lists from those two book is legally possible, we would not provide it, because the lists are the most important intellectual property of their books. As many people are involved in their publication business, we do not want to disrupt their economic activities. If you want to access the list of those two books, please reference the printed books; ISBN codes are available on the reference section.

*D. AI Prediction Accuracy*

After 2023's K-CSAT exam was held on, we compared the words with our AI-Scores. The results are described on Fig 5. The values of K-CSAT explains the appearance rate of words between each AI-Score segment. And the values of Accumulated K-CSAT is the rate of total words, in each segment and higher segment: in short, it's an accumulated rate calculated from the top.

## IV. DISCUSSION

AI-Score works more similar to accumulated K-CSAT distribution rather than K-CSAT result itself. Also, AI-Score worked much better at high-score segments rather than low-score segments. The result showed that ground truth probability of English vocabulary appearance was much higher than AI predicted value.

Therefore, in other words, we assume that any words which AI predicted to be likely to appear on the next exam would appear more often actually. Also, with the comparison result described on Table II, we think that applying human experts' know-how on data driven analysis model would perform much better.

One weak point of our approach was the size of data. As the data appearance pattern between K-CSAT and mock test was too different, we could not directly train our model on mock test to predict K-CSAT patterns. Therefore, if more data were available, much more dramatic result would be expected.

## V. CONCLUSION

We have provided a set of words with high correlation with K-CSAT. We provided preprocessing modules for K-CSAT exam tokenization, and for word appearance tendency analysis. After 2 human experts' screen, 1,300 words remained. Those words set showed much more accuracy rate than previous method s.

Also, we suggested an LSTM based AI model which learns the pattern of vocabulary appearance from K-CSAT exams. It showed higher accuracy on high-probability sections, and lower-probability words tend to appear much more than we expected.

ACKNOWLEDGMENT

We would like to express our great gratitude to Jehun Yoo for OCR-recording of word sheets from *Word Master* and *Kyeong Seon Sik Voca*.

SUPPLEMENT DATA

1. https://github.com/needleworm/bigdata_voca/tree/main/pdf%20parser
2. https://github.com/needleworm/bigdata_voca/blob/main/2023%20sunung%20analysis/extract_sentences_from_sunung.py
3. https://github.com/needleworm/bigdata_voca/blob/main/bigdata%20analysis/train.py
4. https://www.researchgate.net/publication/365678306
5. https://www.researchgate.net/publication/365491164
6. https://www.researchgate.net/publication/365491618
7. https://www.researchgate.net/publication/365670891